\documentclass[11pt,a4paper]{article}
\usepackage[hyperref]{acl2021}
\usepackage{times}
\usepackage{latexsym}

\usepackage{multirow}
\usepackage{graphicx}
\usepackage{subfigure}
\usepackage{enumitem}
\usepackage{color, colortbl}
\usepackage{bm}
\usepackage{amsmath}
\usepackage{amsfonts}
\usepackage{array}
\usepackage{url}
\usepackage{amsthm}
\usepackage{amssymb}
\usepackage{xspace}
\usepackage{xcolor, soul}
\sethlcolor{green}
\definecolor{OliveGreen}{rgb}{0,0.6,0}

\newcommand{\tabincell}[2]{\begin{tabular}{@{}#1@{}}#2\end{tabular}}
\newcolumntype{L}[1]{>{\raggedright\arraybackslash}p{#1}}
\newcolumntype{C}[1]{>{\centering\arraybackslash}p{#1}}
\newcolumntype{R}[1]{>{\raggedleft\arraybackslash}p{#1}}
\definecolor{Gray}{gray}{0.9}

\newcommand{\nop}[1]{}

\definecolor{mypink}{rgb}{0.858, 0.188, 0.478}

\newcommand{\model}{\textsc{QACG}\xspace}
\usepackage{microtype}

\aclfinalcopy % Uncomment this line for the final submission
\def\aclpaperid{1838} %  Enter the acl Paper ID here

\title{Zero-shot Fact Verification by Claim Generation}

\author{Liangming Pan$^{1,2}$ \quad Wenhu Chen$^{3}$ \quad Wenhan Xiong$^{3}$ \\ \textbf{Min-Yen Kan}$^2$ \quad \textbf{William Yang Wang$^{3}$} \\
$^1$NUS Graduate School for Integrative Sciences and Engineering\\
$^2$School of Computing, National University of Singapore, Singapore\\
$^3$University of California, Santa Barbara, CA, USA \\
{\tt liangmingpan@u.nus.edu} \\
{\tt \{wenhuchen, xwhan, william\}@cs.ucsb.edu}\\
{\tt kanmy@comp.nus.edu.sg}\\
}

\date{}

\begin{document}
\maketitle
\begin{abstract}
Neural models for automated fact verification have achieved promising results thanks to the availability of large, human-annotated datasets. However, for each new domain that requires fact verification, creating a dataset by manually writing claims and linking them to their supporting evidence is expensive. We develop \model, a framework for training a robust fact verification model by using automatically-generated claims that can be supported, refuted, or unverifiable from evidence from Wikipedia. \model generates question--answer pairs from the evidence and then convert them into different types of claims. 
Experiments on the FEVER dataset show that our \model framework significantly reduces the demand for human-annotated training data. In a zero-shot scenario, \model improves a RoBERTa model's $F_1$ from 50\% to 77\%, equivalent in performance to 2K+ manually-curated examples. Our \model code is publicly available.\footnote{\url{https://github.com/teacherpeterpan/Zero-shot-Fact-Verification}}

\end{abstract}

\begin{figure*}[!t]
	\centering
	\includegraphics[width=16cm]{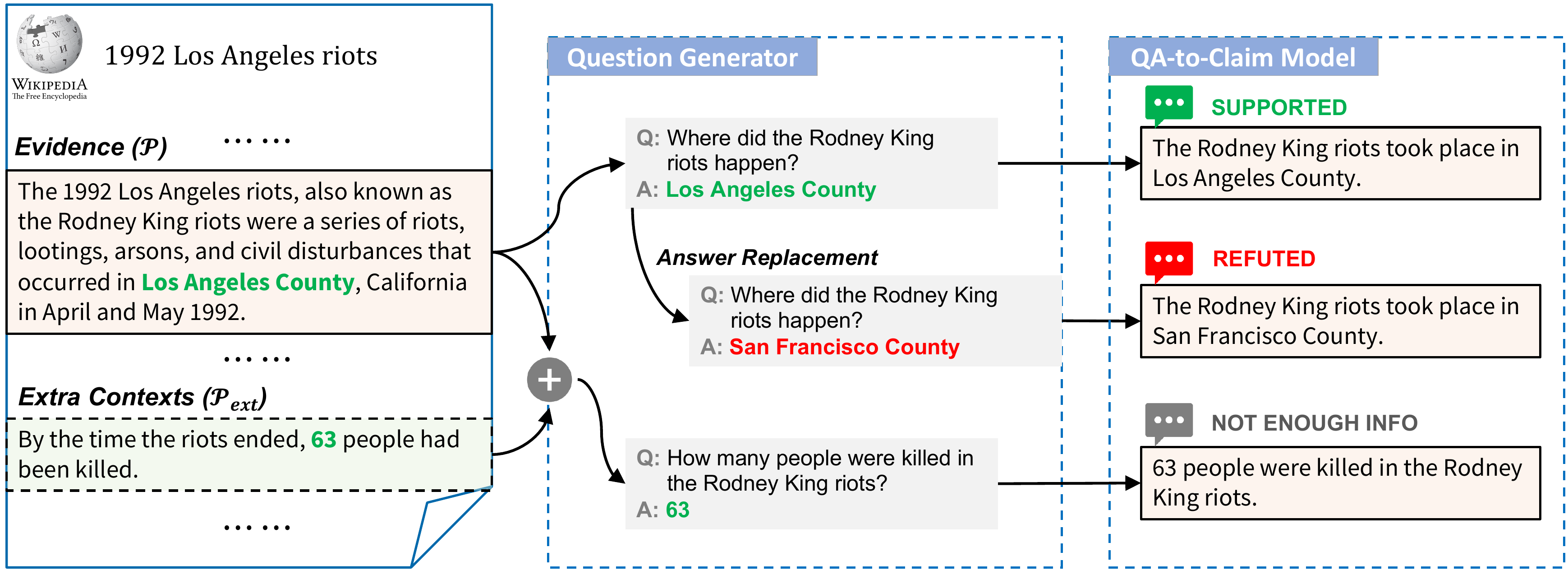}
    \caption{Overview of our \model framework, consisting of two modules: 1) \textit{Question Generator} generates questions from the evidence $\mathcal{P}$ and the extra contexts $\mathcal{P}_{ext}$ given different answers extracted from the passage (in green), and 2) \textit{QA-to-Claim} converts question-answer pairs into claims with different labels. }
    \label{fig:general_framework}
\end{figure*}

\section{Introduction}
% What is fact checking and why it is important?
Fact verification aims to validate a claim in the context of evidence. This task has attracted growing interest with the rise in disinformation in news and social media. Rapid progress has been made by training large neural models~\cite{DBLP:conf/acl/ZhouHYLWLS19,DBLP:conf/acl/LiuXSL20,DBLP:conf/acl/ZhongXTXDZWY20} on the FEVER dataset~\cite{DBLP:conf/naacl/ThorneVCM18}, containing more than 100K human-crafted (evidence, claim) pairs based on Wikipedia. 

% which can either support, refute, or contain not enough information.\ww{Not sure if you need to say the three categories here, because your lines 073-076 are also talking about this. So it's repetitive.} 

% With a rise in deliberate disinformation and hoaxes in propaganda, news, and social media, there has been a growing interest in this task. 

% Why do we need few-shot / zero-shot?
Fact verification is demanded in many domains, including news articles, social media, and scientific documents. However, it is not realistic to assume that large-scale training data is available for every new domain that requires fact verification. Creating training data by asking humans to write claims and search for evidence to support/refute them can be extremely costly. 

% There is an increasing demand for fact verification in many domains
% \xwh{Especially when the human annotators are not familiar with the underlying knowledge resource}
% time-consuming and 

% What is zero-shot fact verification
We address this problem by exploring the possibility of automatically \textit{generating} large-scale (evidence, claim) pairs to train the fact verification model. We propose a simple yet general framework \textbf{Q}uestion \textbf{A}nswering for \textbf{C}laim \textbf{G}eneration (\model) to generate three types of claims from any given evidence: 1) claims that are \texttt{supported} by the evidence, 2) claims that are \texttt{refuted} by the evidence, and 3) claims that the evidence does Not have Enough Information (\texttt{NEI}) to verify. 

% to ask different questions for given evidence and convert question-answer pairs into claims with different labels. Question generation

% What is QG? Why choosing QG?
To generate claims, we utilize \textit{Question Generation (QG)} ~\cite{DBLP:conf/emnlp/ZhaoNDK18,DBLP:conf/www/LiuBang20,DBLP:conf/acl/PanXFCK20}, which aims to automatically ask questions from textual inputs. QG has been shown to benefit various NLP tasks, such as enriching QA corpora~\cite{DBLP:conf/acl/AlbertiAPDC19}, checking factual consistency for summarization~\cite{DBLP:conf/acl/WangCL20}, and data augmentation for semantic parsing~\cite{DBLP:conf/acl/BansalPG18}. To the best of our knowledge, we are the first to employ QG for fact verification.

% Details about our model
As illustrated in Figure~\ref{fig:general_framework}, given a passage $P$ as the evidence, we first employ a \textit{Question Generator} to generate a question--answer pair $(Q, A)$ for the evidence. We then convert $(Q, A)$ into a claim $C$ (\textit{QA-to-Claim}) based on the following logical assumptions: a) if $P$ can answer $Q$ and $A$ is the correct answer, then $C$ is a \texttt{supported} claim; b) if $P$ can answer $Q$ but $A$ is an incorrect answer, then $C$ is a \texttt{refuted} claim; c) if $P$ cannot answer $Q$, then $C$ is a \texttt{NEI} claim. The Question Generator and the QA-to-Claim model are off-the-shelf BART models~\cite{DBLP:conf/acl/LewisLGGMLSZ20}, finetuned on SQuAD~\cite{DBLP:conf/emnlp/RajpurkarZLL16} and  QA2D~\cite{DBLP:journals/corr/abs-1809-02922} datasets. 

% Evaluation Results
We generate 100K (evidence, claim) pairs for each type of claim, which we then use to train a RoBERTa~\cite{DBLP:journals/corr/abs-1907-11692} model for fact verification. We evaluate the model on three test sets based on the FEVER dataset. Although we do not use any human-labeled training examples, the model achieves over 70\% of the $F_1$ performance of a fully-supervised setting. By finetuning the model with only 100 labeled examples, we further close the performance gap, achieving 89.1\% of fully-supervised performance. The above results show that pretraining the fact verification model with generated claims greatly reduces the demand for in-domain human annotation. When evaluating the model on an unbiased test set for FEVER, we find that training with generated claims also produces a more \textit{robust} fact verification model. 

In summary, our contributions are: 

\noindent $\bullet$ To the best of our knowledge, this is the first work to investigate zero-shot fact verification. 

\noindent $\bullet$ We propose \model, a novel framework to generate high-quality claims via question generation. 

\noindent $\bullet$ We show that the generated training data can greatly benefit the fact verification system in both zero-shot and few-shot learning settings. 

% Min: this can be pushed to the evaluation or bullet point.  Omitted.
% We then evaluate the model on three test sets based on the FEVER dataset and have three major findings. 
% \ww{The bullet points are way too long. I'm not sure if you need to discuss detailed results in the introduction: normally we don't. The three contributions have to be concise. Maybe a short paragraph to summarize your results, and then three short bullet points for your contributions.} \kmy{Agreed.  This should introduce rather than summarize.  Make sure you get the reader to read in the other sections later; don't be repetitive.} \kmy{Also, what is the main reason that this paper should be accepted?  It seems like you are relying on performance figures rather than ingenuity for acceptance; all your claims are tied to performance.  Shouldn't they be tied to interesting methods instead?  Is it the claim generation or the overall 2-module framework that is interesting?  How does it differ from other uses of QG for enhancing performance (e.g., QA performance)?}

\section{Methodology}
\label{sec:method}

% \subsection{Problem Formulation}

Given a claim $\mathcal{C}$ and a piece of evidence $\mathcal{P}$ as inputs, a \textit{fact verification} model $\mathcal{F}$ predicts a label $\mathcal{Y} \in \{ \texttt{supported}, \texttt{refuted}, \texttt{NEI} \}$ to verify whether $\mathcal{C}$ is supported, refuted, or can not be verified by the information in $\mathcal{P}$. 

For the \textit{zero-shot} setting, we assume no human-annotated training example is available. Instead, we generate a synthetic training set based on our \model framework to train the model. 

\subsection{Question Generator and QA-to-Claim}

As illustrated in Figure~\ref{fig:general_framework}, our claim generation model \model has two major components: a \textit{Question Generator} $\mathcal{G}$, and a \textit{QA-to-Claim} model $\mathcal{M}$. 

The \textbf{Question Generator} takes as input an evidence $\mathcal{P}$ and a text span $A$ from the given evidence and aims to generate a question $Q$ with $A$ as the answer. 
% To implement this, we use the SQuAD data split from~\citet{DBLP:conf/nlpcc/ZhouYWTBZ17} to fine-tune the Google T5 model~\cite{DBLP:journals/jmlr/RaffelSRLNMZLL20}, where the model encodes the concatenation of the SQuAD passage and answer and learns to decode the question.  
We implement this with the BART model~\cite{DBLP:conf/acl/LewisLGGMLSZ20}, a large transformer-based sequence-to-sequence model pretrained on 160GB of text. The model is finetuned on the SQuAD dataset processed by~\citet{DBLP:conf/nlpcc/ZhouYWTBZ17}, where the model encodes the concatenation of the SQuAD passage and the answer text and then learns to decode the question. We evaluate the question generator using automatic and human evaluation and investigate its impact on fact verification in Appendix~\ref{app:qg_evaluation}. 

The \textbf{QA-to-Claim Model} takes as inputs 
% a question 
$Q$ and % an answer 
$A$, and outputs the declarative sentence $C$ for the $(Q, A)$ pair, as shown in Figure~\ref{fig:general_framework}. 
% For example, given $Q = \texttt{Who called Taylor?}$ and $A = \texttt{Liz}$, the output $C = \texttt{Liz called Taylor}$. 
We also treat this as a sequence-to-sequence problem and finetune the BART~\cite{DBLP:conf/acl/LewisLGGMLSZ20} model on the QA2D dataset~\cite{DBLP:journals/corr/abs-1809-02922}, which contains the human-annotated declarative sentence for each $(Q, A)$ pair in SQuAD. 

\subsection{Claim Generation}

Given the pretrained question generator $\mathcal{G}$ and the QA-to-Claim model $\mathcal{M}$, we then formally introduce how we generate claims with different labels. 

\paragraph{\texttt{Supported} claim generation.}
Given an evidence $P$, we use named entity recognition to identify all entities within $P$, denoted as $\mathcal{E}$. For each entity $a \in \mathcal{E}$, we treat each $a$ in turn as an answer and generate a question $q = \mathcal{G}(\mathcal{P}, a)$ with the question generator. The question--answer pair $(q, a)$ are then sent to the QA-to-Claim model to generate the \texttt{supported} claim $c = \mathcal{M}(q, a)$.

\paragraph{\texttt{Refuted} claim generation.}
To generate a \texttt{refuted} claim, after we generate the question--answer pair $(q, a)$, we use \textit{answer replacement} (shown in Figure~\ref{fig:general_framework}) to replace the answer $a$ with another entity $a'$ with the same type such that $a'$ becomes an incorrect answer to the question $q$. Using $a$ as the query, we randomly sample a phrase from the top-$5$ most similar phrases in the pretrained Sense2Vec~\cite{DBLP:journals/corr/TraskML15} as the replacing answer $a'$. The new pair $(q, a')$ is then fed to the QA-to-Claim model to generate the \texttt{refuted} claim. 

% Add discussion about limitations here...
% (for example,  $a$ is ``James Cameron'', and $a'$ is ``James Francis Cameron'')
To avoid the case that $a'$ is still the correct answer, we define rules to ensure that the $a'$ has less lexical overlap with $a$. 
% However, there remain a few cases that are difficult to address. 
However, this problem is sometimes non-trivial and cannot be completely avoided. For example, for the QA pair: (``Who is the producer of Avatar?''; ``James Cameron''), another valid answer $a'$ is ``Jon Landau'', who happens to be another producer of Avatar. % It requires world knowledge to avoid $a'$ being ``Jon Landau''. 
However, we observe that such coincidences rarely happen: among the 100 randomly sampled claims, we only observed 2 such cases. 
Therefore, we leave them as the natural noise of the generation model. 

\paragraph{\texttt{NEI} claim generation.}
% To generate an \texttt{NEI} claim from the evidence $\mathcal{P}$, 
We need to generate a question $q'$ which is relevant but cannot be answered by $\mathcal{P}$. To this end, we link $\mathcal{P}$ back to its original Wikipedia article $\mathcal{W}$ and expand the evidence with additional contexts $\mathcal{P}_{ext}$, which are five randomly-retrieved sentences from $\mathcal{W}$ that are not present in $\mathcal{P}$. In our example in Figure~\ref{fig:general_framework}, one additional context retrieved is ``By the time the riots ended, 63 people had been killed''. We then concatenate $\mathcal{P}$ and $\mathcal{P}_{ext}$ as the expanded evidence, based on which we generate a \texttt{supported} claim given an entity in $\mathcal{P}_{ext}$ as the answer (\textit{e.g.}, ``63''). This results in a claim relevant to but unverifiable by the original evidence $\mathcal{P}$. 

\section{Experiments}

By applying our \model model to each of the $18,541$ Wikipedia articles in the FEVER training set, we generate a total number of $176,370$ \texttt{supported} claims, $360,924$ \texttt{refuted} claims, and $258,452$ \texttt{NEI} claims. Our generated data is around five times the size of the human-annotated claims in FEVER. We name this generated dataset as \textit{\model-Full}. We then create a balanced dataset \textit{\model-Filtered} by randomly sampling $100,000$ samples for each class. Statistics of the FEVER and the generated dataset are in Appendix~\ref{app:data_statistics}. % Examples of generated claims are presented in Table~\ref{tbl:claim_examples}. 

\begin{table*}[!t]
    \small
	\begin{center}
	\renewcommand{\arraystretch}{1.1}
		\begin{tabular}{ c  l | c | c | c } \hline
        \multicolumn{2}{c|}{\multirow{2}{*}{Model}} & \textbf{\tabincell{c}{FEVER\\-Symmetric}} & \textbf{FEVER-S/R} & \textbf{FEVER-S/R/N} \\ \cline{3-5}
        & & $P$ / $R$ / $F_1$ & $P$ / $R$ / $F_1$ & $P$ / $R$ / $F_1$ \\ \hline \hline
        % Supervised
        \multirow{2}{*}{\textit{Supervised}} & S1. BERT-base~\cite{DBLP:conf/naacl/DevlinCLT19} & 81.5 / 81.3 / 81.2 & 92.8 / 92.6 / 92.6 & 85.7 / 85.6 / 85.6 \\
        & S2. RoBERTa-large~\cite{DBLP:journals/corr/abs-1907-11692} & \textbf{85.5 / 85.5 / 85.5} & \textbf{95.2 / 95.1 / 95.1} & \textbf{88.0 / 87.9 / 87.8}  \\ \hline
        % Zero-shot
        \multirow{6}{*}{\textit{Zero-shot}} & U1. Random Guess & 50.0 / 50.0 / 50.0 & 50.0 / 50.0 / 50.0 & 33.3 / 33.3 / 33.3 \\
        & U2. GPT2 Perplexity & 52.7 / 52.7 / 52.7 & 55.6 / 55.6 / 55.6 & 35.3 / 35.3 / 35.3 \\
        & U3. MNLI-Transfer & 62.2 / 55.5 / 58.7 & 63.6 / 60.5 / 61.8 & 41.4 / 39.6 / 40.7 \\
        & U4. LM as Fact Checker~\cite{DBLP:LM_as_fact_checker} & 71.2 / 64.5 / 67.8 & 77.9 / 65.6 / 70.2 & 64.3 / 54.6 / 49.8 \\
        % & U2. Claim Extractor &  &  &  \\ 
        & U5. \model (BERT-base) & 73.2 / 73.0 / 72.9 & 74.2 / 74.0 / 74.1 & 56.5 / 55.7 / 55.9 \\ 
        & U6. \model (RoBERTa-large) & \textbf{77.3 / 77.0 / 77.1} & \textbf{78.1 / 78.1 / 78.1} & \textbf{64.6 / 62.0 / 62.6} \\ \hline
        % Few-shot
        % \multirow{3}{*}{\textit{Few-Shot}} & E1. RoBERTa-large & 64.1 / 62.8 / 63.3 & 70.6 / 69.7 / 69.4 & 64.4 / 63.1 / 63.5 \\
        % & E2. MNLI-Finetune & 72.5 / 72.5 / 72.5 & 86.4 / 86.2 / 86.3 & 72.5 / 72.3 / 72.4 \\ 
        % & E3. \model + RoBERTa-large & \textbf{80.9 / 80.8 / 80.8} & \textbf{89.7 / 89.7 / 89.7} & \textbf{78.1 / 78.2 / 78.1} \\ \hline
		\end{tabular}
	\end{center}
\vspace{-0.2cm}
\caption{Fact verification performance for supervised models and zero-shot models on three different settings. }
\vspace{-0.2cm}
\label{tbl:zeroshot_performance}
\end{table*}

\paragraph{Evaluation Datasets.} 
We evaluate fact verification on three different test sets based on FEVER: \textbf{1) FEVER-S/R}: Since only the \texttt{supported} and \texttt{refuted} claims are labeled with gold evidence in FEVER, we take the claim--evidence pairs of these two classes from the FEVER test set for evaluation. \textbf{2) FEVER-Symmetric}: this is a carefully-designed unbiased test set designed by~\citet{DBLP:conf/emnlp/SchusterSYFSB19} to detect the robustness of the fact verification model. Note that only \texttt{supported} and \texttt{refuted} claims are present in this test set. \textbf{3) FEVER-S/R/N}: The full FEVER test set are used for a three-class verification. We follow~\citet{DBLP:conf/emnlp/AtanasovaWA20} to use the system of~\citet{DBLP:journals/corr/abs-1901-02534} to retrieve evidence sentences for \texttt{NEI} claims. 
% MinCR: need details to make this more clear how the baseline Malon 2019 system works?  Otherwise impossible for the reader to make sense of the method.

\paragraph{Fact Verification Models.} As shown in Table~\ref{tbl:zeroshot_performance}, we take a BERT model (S1) and a RoBERTa model (S2) fine-tuned on the FEVER training set as the \textit{supervised} models. Their corresponding \textit{zero-shot} settings are Rows~U5 and U6, where the models are trained on our generated \textit{\model-Filtered} dataset. Note that for binary classification (FEVER-S/R and FEVER-Symmetric), only the \texttt{supported} and \texttt{refuted} claims are used for training, while for FEVER-S/R/N, the full training set is used. 

We employ four baselines that also do not need any human-annotated claims to compare with our method. \textit{Random Guess} (U1) is a weak baseline that randomly predicts the class label. \textit{GPT2 Perplexity} (U2) predicts the class label based on the perplexity of the claim under a pretrained GPT2~\cite{radford2019language} language model, following the assumption that ``misinformation has high perplexity''~\cite{DBLP:journals/corr/abs-2006-04666}. \textit{MNLI-Transfer} (U3) trains a BERT model for natural language inference on the MultiNLI corpus~\cite{DBLP:conf/naacl/WilliamsNB18} and applies it for fact verification. \textit{LM as Fact Checker}~\cite{DBLP:LM_as_fact_checker} (U4) leverages the implicit knowledge stored in the pretrained BERT language model to verify a claim. The implementation details are given in Appendix~\ref{app:baseline_implementation}. 
% to create an effective end-to-end fact checker using a solely a language model,

\subsection{Main Results}
% \kmy{It might be more succinct to state up front you are going to measure performance against the fully-supervised model as a performance gap (the lower the better).  You can use the F1 scores as ancillary results but not worth mentioning normally.  That seems to be the big part of your argument anyways.}
Table~\ref{tbl:zeroshot_performance} summarizes the fact verification performance, measured by the macro Precision ($P$), Recall ($R$), and F1 Score ($F_1$). 

% U3 performs good compared with S2
% RoBERTa performs better than BERT
\paragraph{Comparison with supervised settings. }
The zero-shot setting with RoBERTa-large (U6) attains 78.1 $F_1$ on the FEVER-S/R and 62.6 $F_1$ on the FEVER-S/R/N. The $F_1$ gap to the fully-supervised RoBERTa-large (S2) is only $17.0$ and $15.2$ on these two settings, respectively. These results demonstrate the effectiveness of \model in generating good (evidence, claim) pairs for training the fact verification model. The RoBERTa model (S2, U6) is more effective than the BERT model (S1, U5) for both the zero-shot and the supervised setting. 

\paragraph{Comparison with zero-shot baselines. }
Our model (U6) achieves the best results among all the zero-shot baselines across all three test sets. We find that validating a claim by its perplexity (U2) only works slightly better than random guess (U1) (+3.43 $F_1$), showing that misinformation does not necessary to have high perplexity. Although natural language inference seems highly correlated with fact verification, directly transferring the model trained on the MNLI dataset (U3) only outperforms random guess by 9.30 $F_1$. We believe this is due to the domain gap between FEVER (from Wikipedia) and the MNLI (from fiction, letters, etc.) dataset. 
% Our model avoids this issue since we can \textit{generate} synthetic training data from the same domain. 
As a generation framework, our model can avoid the domain gap issue by generating pseudo training data from the same domain (Wikipedia). Another reason is the ``task gap'' between NLI and fact verification, in which the former makes inference about the situation described in a sentence, while the latter focuses on claims about entities in Wikipedia. 
% The kinds of data you’ll get for these two tasks are going to be quite different even if they are annotated in the same domain.

% Less performance drop when transferred to FEVER symmetric
\paragraph{Model Robustness.}
We observe a large performance drop when the supervised model is evaluated on the FEVER-Symmetric test set for both the BERT model ($-11.4$ $F_1$) and the RoBERTa model ($-9.6$ $F_1$). However, the models trained with our generated data (U2, U3) drop only $1.2$ and $1.0$ $F_1$ drop. This suggests that the wide range of different claims we generate as training data helps eliminate some of the annotation artifacts present in FEVER, leading to a more robust fact verification model. 

\begin{figure}[!t]
	\centering
	\includegraphics[width=7.5cm]{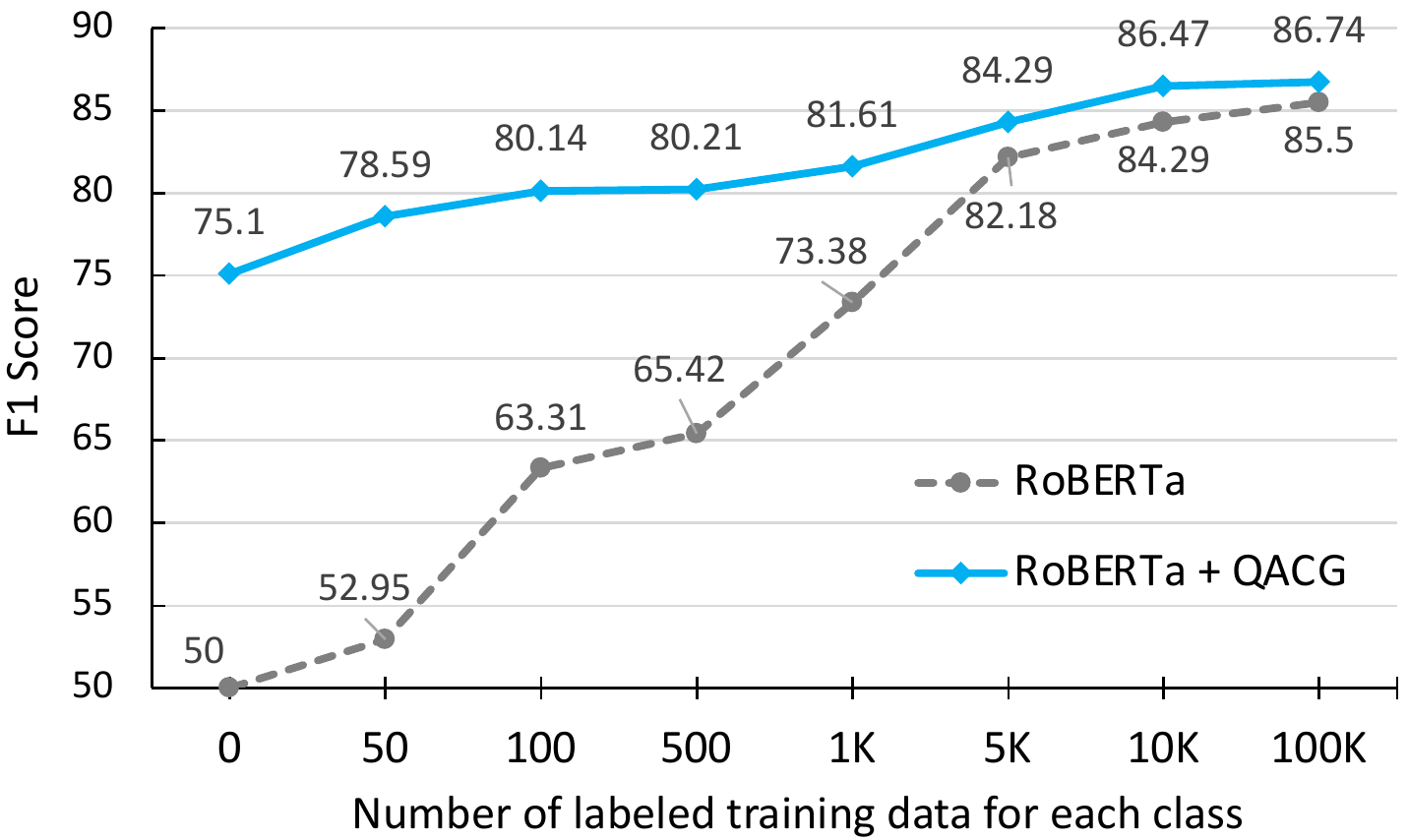}
	% \vspace{-0.3cm}
    \caption{The few-shot learning experiment. The figure shows the $F_1$ score on FEVER-Symmetric for progressively larger training dataset sizes. }
    \vspace{-0.3cm}
    \label{fig:few_shot_Symmetric}
\end{figure}

\begin{table*}[!t]
    \small
	\begin{center}
	    \renewcommand{\arraystretch}{1.1}
		\begin{tabular}{ l l } \hline
        \textbf{Evidence} & \textbf{Generated Claim}  \\ \hline
        % Example 1
        \multirow{9}{*}{\tabincell{l}{{\color{blue} Budapest} is cited as one of the most beautiful \\ cities in {\color{blue} Europe}, ranked as the most liveable \\ Central and {\color{blue} Eastern European} city on {\color{blue} EIU}'s \\ quality of life index, ranked as ``the world's \\ {\color{blue} second} best city'' by {\color{blue} Conde Nast Traveler}, \\ and ``{\color{blue} Europe}'s {\color{blue} 7th} most idyllic place to live'' \\ by {\color{blue} Forbes}.}} & \textbf{\texttt{SUPPORTED} claims} \\ 
        & $\quad$ {\color{blue} Budapest} is ranked as the most liveable city in central Europe. \\
        & $\quad$ Budapest ranks {\color{blue} 7th} in terms of idyllic places to live in Europe. \\
        & \textbf{\texttt{REFUTED} claims} \\
        & $\quad$ Budapest ranks in {\color{red} 11th} in terms of idyllic places to live in Europe. \\
        & $\quad$ Budapest is ranked the most liveable city in {\color{red} Asia}. \\
        & \textbf{\texttt{NEI} claims} \\
        & $\quad$ Budapest is one of the largest cities in the European Union. \\
        & $\quad$ Budapest is the capital of Hungary. \\ \hline
        % Example 2
        \multirow{9}{*}{\tabincell{l}{{\color{blue} Alia Bhatt} received critical acclaim for portraying \\ emotionally intense characters in the road drama \\ {\color{blue} Highway (2014)}, which won her the {\color{blue} Filmfare} \\ {\color{blue} Critics Award} for {\color{blue} Best Actress}, and the crime \\ drama {\color{blue} Udta Punjab (2016)}, which won her the \\ {\color{blue} Filmfare Award for Best Actress.} }} & \textbf{\texttt{SUPPORTED} claims} \\ 
        & $\quad$ Bhatt won the {\color{blue} Filmfare Award for Best Actress} in Udta Punjab. \\
        & $\quad$ Bhatt received the Filmfare Critics Award for her role in {\color{blue} Highway}. \\
        & \textbf{\texttt{REFUTED} claims} \\
        & $\quad$ Alia Bhatt won the {\color{red} Best Original Screenplay} award in Highway. \\
        & $\quad$ {\color{red} 2 States (2014)} won Alia Bhatt the Filmfare Award for Best Actress. \\
        & \textbf{\texttt{NEI} claims} \\
        & $\quad$ Alia Bhatt made her acting debut in the 1999 thriller Sangharsh. \\
        & $\quad$ Bhatt played her first leading role in Karan Johar's romantic drama. \\ \hline
		\end{tabular}
	\end{center}
\vspace{-0.2cm}
\caption{Examples of evidence and claims generated by \model, categorized by class labels. In the evidence, the identified answers for question generation are highlighted in blue. For claims, the correct answers are highlighted in blue for \texttt{SUPPORTED} claims and the replaced wrong answers are in red for \texttt{REFUTED} claims. }
\vspace{-0.2cm}
\label{tbl:claim_examples}
\end{table*}

\begin{table}[!t]
    \small
	\begin{center}
	    \renewcommand{\arraystretch}{1.1}
		\begin{tabular}{ l l } \hline
        \textbf{Evidence:} & Roman Atwood is \underline{best known for his vlogs,} \\
        & \underline{where he posts updates about his life.} \\
        \textbf{Claim:} & \tabincell{l}{Roman Atwood is \underline{a content creator.}} \\ \hline
        \textbf{Evidence:} & In 2004, Slovenia \underline{entered NATO and the} \\
        & \underline{European Union. } \\
        \textbf{Claim:} & \tabincell{l}{Slovenia \underline{uses the euro.}} \\ \hline
        \textbf{Evidence:} & He has traveled to \underline{Chad and Uganda} to raise \\
        & awareness about conflicts in the regions. \\
        \textbf{Claim:} & \tabincell{l}{Ryan Gosling has been to \underline{a country in Africa.}} \\ \hline
		\end{tabular}
	\end{center}
\vspace{-0.2cm}
\caption{Examples of claims in FEVER that require commonsense or world knowledge (underlined). }
\vspace{-0.2cm}
\label{tbl:reasoning_claim_examples}
\end{table}

\subsection{Few-shot Fact Verification}

% The \textit{few-shot} section in Table~\ref{tbl:zeroshot_performance} shows the few-shot performance where only a few human-labeled (evidence, claim) pairs are available. For our model (E3), we first pretrain the RoBERT-large with our generated dataset \model-Filtered and then fine-tune the model with $100$ human-labeled claims in FEVER. We compare this with the model that trains with the $100$ claims from scratch (E1) and the model pretrained on the MNLI dataset (E2). 

% We find that our model significantly outperforms the two baselines across all three settings. To better demonstrate this, in Figure~\ref{fig:few_shot}, we plot the performance of E1 (grey curves) and E3 (blue curves) with progressively larger sizes of human-labeled training data. We find that our model performs consistently better than the model without pretraining with generated data. The performance improvement is especially prominent in very data-poor regimes; for example, our approach achieves $75.2$ F1 with only 50 labeled claims for each class on FEVER-S/R/N, compared with $46.8$ F1 without pretraining ($28.4$ absolute gain). 

% \kmy{I thought few shot is usually less than 10?  Call this limited training data?}
We then explore \model's effectiveness in the few-shot learning setting where only a few human-labeled (evidence, claim) pairs are available. We first train the RoBERT-large fact verification model with our generated dataset \model-Filtered. Then we fine-tune the model with a limited amount of human-labeled claims in FEVER. The blue solid line in Figure~\ref{fig:few_shot_Symmetric} shows the $F_1$ scores on FEVER-Symmetric after finetuning with different numbers of labeled training data. We compare this with training the model from scratch with the human-labeled data (grey dashed line). % The F1 curves for the other two test sets are in Appendix E, showing similar results. 

% With progressively larger training dataset sizes, 
Our model performs consistently better than the model without pretraining, regardless of the amount of labeled training data. The improvement is especially prominent in data-poor regimes; for example, our approach achieves $78.6$ $F_1$ with only 50 labeled claims for each class, compared with $52.9$ $F_1$ without pretraining ($+25.7$). This only leaves a $7.9$ $F_1$ gap to the fully-supervised setting ($86.5$ $F_1$) with over 100K training samples. 
% MinCR: then the question is what types of information are in the 100k that are missing from your cases?  Is it because of the form of the question?  The vocabulary mismatch?  What?
The results show pretraining fact verification with \model greatly reduces the demand for in-domain human-annotated data. Our method can provide a ``warm start'' for fact verification system when applied to a
% MinCR: not sure I 100% agree with this.  The system works because you are testing on data similar to the source of the augmentation (Wikipedia), right?  Does this really mean you have a new domain?
new domain where training data are limited. 

% Moreover, pretraining with generated data and finetuning with a few annotated data can attain a very close performance to the fully-supervised setting. For example, $89.25$ F1 is achieved after finetuing with $50$ examples for each class, which only leaves a $6.13$ F1 gap to the fully-supervised setting with over 100K training samples. The results show pretraining fact verification with \model greatly reduce the demand for in-domain human-annotated data. Our method can provide a ``warm start'' for fact verification system when it is applied to a new domain where training data are quite limited. 

\subsection{Analysis of Generated Claims}
\label{app:claim_example}

% Short Version
% Table~\ref{tbl:claim_examples} shows representative claims generated by our model. The claims are fluent, label-cohesive, and exhibit encouraging language variety. However, one limitation is that most of our generated claims do not perform deep reasoning over the evidence, since the question generator is finetuned on the SQuAD dataset, in which more than 80\% of its questions are shallow factoid~\cite{DBLP:conf/acl/DuSC17}. 

% To better understand the impact of this limitation, we randomly sampled 200 claims from the FEVER dataset. We find that 38\% of the \texttt{supported} claims and 16\% of the \texttt{refuted} claims require either commonsense reasoning or world knowledge to verify. Table~\ref{tbl:reasoning_claim_examples} show three typical examples. Therefore, we believe that this limitation brings a domain gap between the generated claims and the human-written claims, which is the main bottleneck of our system before it can be applied to more complex domains for fact verification. 

% Long Version
Table~\ref{tbl:claim_examples} shows representative claims generated by our model. The claims are fluent, label-cohesive, and exhibit encouraging language variety. However, one limitation is that our generated claims are mostly \textit{lack of deep reasoning over the evidence}. This is because we finetune the question generator on the SQuAD dataset, in which more than 80\% of its questions are shallow factoid questions. 
% ~\cite{DBLP:conf/acl/DuSC17}. 

% MinCR: not clear about what you mean as a domain gap?  You just mean a difference in performance due to nature of the difference between training and testing data?  @abhinav also critiques other's for misuse of the term "domain", you may want to discuss with him over Slack.
To better understand whether this limitation brings a domain gap between the generated claims and the human-written claims, we randomly sampled 100 \texttt{supported} claims and 100 \texttt{refuted} and analyze whether reasoning is involved to verify those claims. We find that 38\% of the \texttt{supported} claims and 16\% of the \texttt{refuted} claims in FEVER require either commonsense reasoning or world knowledge to verify. Table~\ref{tbl:reasoning_claim_examples} show three typical examples. 
% MinCR: need to narrate / explain the reasoning here for those examples.
Therefore, we believe this domain gap is the main bottleneck of our system. Future studies are required to generate more complex claims which involves multi-hop, numerical, and commonsense reasoning, such that we can apply our model to more complex fact checking scenario. 
% fact verification to more complex domains without much human annotation. 

% \section{Related Works}

% \paragraph{Fact Checking} involves assessing the truthfulness of a claim, 

% Fact checking systems consist of components to identify check-worthy claims (Atanasova et al., 2018; Hansen et al., 2019; Wright and Augenstein, 2020), retrieve and rank evidence documents (Yin and Roth, 2018; Allein et al., 2020), determine the relationship between claims and evidence doc- uments (Bowman et al., 2015; Augenstein et al., 2016; Baly et al., 2018), and finally predict the claims’ veracity (Thorne et al., 2018; Augenstein et al., 2019). 

% \paragraph{Question Generation}

% useful in many cases
% pretraining QUESUS
% factual consistency: summarization
% we are the first to link xxx 

\section{Conclusion and Future Work}
We utilize the question generation model to ask different questions for given evidence and convert question--answer pairs into claims with different labels. We show that the generated claims can train a well-performing fact verification model in both the zero-shot and the few-shot learning setting. Potential future directions could be: 1) generating more complex claims that require deep reasoning; 2) extending our framework to other fact checking domains beyond Wikipedia, \textit{e.g.}, news, social media; 3) leveraging generated claims to improve the robustness of fact checking systems. 

% real, low-resource scenarios to prove our approach’s robustness. 

% Further work is required to generate more complex claims that involves multi-hop, numerical, and commonsense reasoning, so that we can apply fact verification to more complex domains without much human annotation. 

\section*{Acknowledgments}

This research is supported by the National Research Foundation, Singapore under its International Research Centres in Singapore Funding Initiative. 
% Any opinions, findings and conclusions or recommendations expressed in this material are those of the author(s) and do not reflect the views of National Research Foundation, Singapore. 
% UCSB authors are not supported by any of the above projects. 
The UCSB authors are not supported by any of the projects above. They thank Google, Amazon, Facebook, and JP Morgan for their generous support. 

\section*{Ethical Considerations}

% The core technique we propose is claim generation, which aims to automatically generate claims that are either supported, refuted, or not verifiable by given evidence. In this section, 

We discuss two potential issues of claim generation, showing how our work sidesteps these issues. 
While individuals may express harmful or biased claims, our work only focuses on generating factoid claims from a corpus.  In this work, we take Wikipedia as the source for objective fact.  Practicing this technique thus requires the identification of an appropriate source of objective truth to generate claims from. 
Another potential misuse of claim generation is to generate \texttt{refuted} claims and subsequently spread such misinformation.  We caution practitioners to treat the generated claims with care.  In our case, we use the generated claims only to optimize for the downstream fact verification task.  We advise against releasing generated claims for public use --- especially on public websites, where they may be crawled and then subsequently used for inference.  As such, we will release the model code but not the output in our work.  Practitioners can re-run the training pipeline to replicate experiments accordingly.

% Although the generated claims are used to optimize for downstream QA performance, it is still instructive to examine the generated claims to better understand our system's advantages and limitations. 

\bibliographystyle{acl_natbib}
% \bibliography{anthology,acl2021}
\bibliography{acl2021}

% \clearpage

% {\Large \noindent \textbf{Supplementary Materials}}

\appendix

\section{Evaluation of Question Generation}
\label{app:qg_evaluation}

To implement the question generator, we finetune the pretrained BART model provided by HuggingFace library on the SQuAD dataset. The codes are based on the SimpleTransformers\footnote{https://github.com/ThilinaRajapakse/simpletransformers} library. The success of our \model framework heavily rely on whether we can generate fluent and answerable questions given the evidence. Therefore, we separately evaluate the question generator using both automatic and human evaluation and investigate its impact to zero-shot fact verification. 

\subsection{Automatic Evaluation}

We employ BLEU-4~\cite{DBLP:conf/acl/PapineniRWZ02}, METEOR~\cite{DBLP:conf/wmt/LavieA07}, and ROUGE-L~\cite{lin2004rouge} to evaluate the performance of our implementation. We compare the BART model with several state-of-the-art QG models, using their reported performance on the Zhou split of SQuAD. 

Table~\ref{tbl:qg_evaluation} shows the evaluation results comparing against all baseline methods. The BART model achieves a BLEU-4 of 21.32, outperforming NQG++, S2ga-mp-gsa, and CGC-QG by large margins. This is as expected since these three baselines are based on Seq2Seq and do not apply language model pretraining. Compared with the current state-of-the-art model UniLM, the BART model achieves comparable results, with slightly lower BLEU-4 but higher METEOR.

\begin{table}[htb]
  \small
  \begin{center}
      \begin{tabular}{l|ccc}
      \hline
        Model & B4 & MR & $R_L$ \\ \hline \hline
        NQG++~\cite{DBLP:conf/nlpcc/ZhouYWTBZ17} & 13.5 & 18.2 & 41.6 \\ 
        S2ga-mp-gsa~\cite{DBLP:conf/emnlp/ZhaoNDK18} & 15.8 & 19.7 & 44.2 \\ 
        CGC-QG~\cite{DBLP:conf/www/LiuBang20} & 17.6 & 21.2 & 44.5 \\
        UniLM~\cite{DBLP:conf/nips/00040WWLWGZH19} & \textbf{23.8} & 25.6 & \textbf{52.0} \\ 
        BART~\cite{DBLP:conf/acl/LewisLGGMLSZ20} & 21.3 & \textbf{27.1} & 43.6 \\\hline
      \end{tabular}
  \end{center}
  \caption{Performance evaluation of the \textit{Question Generator} with different model implementations. We adopt the BART model in our \model framework. \textit{B4}: BLEU-4, \textit{MR}: METEOR, $R_L$: ROUGE-L. }
  \label{tbl:qg_evaluation}
  % \vspace{-0.3cm}
\end{table}

\begin{table}[htb]
  \small
  \begin{center}
      \begin{tabular}{l|cc}
      \hline
        Model & Answerable & FV Performance \\ 
        & Rate & $P$ / $R$ / $F_1$ \\ \hline \hline
        NQG++ & 63.0\% & 62.2 / 62.4 / 62.3 \\ 
        BART & 89.5\% & 76.3 / 76.0 / 76.1 \\\hline
      \end{tabular}
  \end{center}
  \caption{\textit{Answerable Rate}: the ratio of answerable questions generated by the NQG++ and the BART model. \textit{FV Performance}: the zero-shot fact verification performance on the FEVER-Symmetric.}
  \label{tbl:answer_evaluation}
\end{table}

\subsection{Impact of Answerability}

Given the evidence $P$ and the answer $A$, the generated question $Q$ must be answerable by $P$ and take $A$ as its correct answer. This is the premise of generating a correct SUPPORTED claim. Therefore, we specially evaluate this \textit{answerability} property via human ratings. We randomly sample 100 generated question-answer pairs with their corresponding evidence and ask two workers to judge the answerability of each sample. We do this for both the NQG++ model and the BART model. To investigate the impact of question quality on the fact verification performance, we separately use the NQG++ and BART as the question generator to generate claims and train the RoBERTa model. The performance is summarized in Table~\ref{tbl:answer_evaluation}. 

We find that the ratio of answerable questions generated by the BART model is $89.5\%$, significantly outperforms the $63.5\%$ achieved by the NQG++ model. When switching the question generator to NQG++, the fact verification $F_1$ drops to $62.3$ ($-22.1\%$ compared with BART). This shows that answerability plays an important role in ensuring the validity of the generated claims and has a huge impact on the fact verification performance. 

\section{Dataset Statistics}
\label{app:data_statistics}

Table~\ref{tbl:datasets} shows the basic data statistics of the FEVER, FEVER-Symmetric, and our generated dataset by \model. We use the balanced dataset \textit{\model-Filtered} sampled from \textit{\model-Full} to train the fact verification model in the zero/few-shot setting. Compared with the original FEVER dataset, our generated \textit{\model-Filtered} dataset has a balanced number of claims for each class. Moreover, because \model can generate three different types of claims for the same given evidence (shown in Figure~\ref{fig:general_framework}), it results in a more ``unbiased'' dataset in which the model must rely on the \textit{(evidence, claim)} pair rather than the \textit{evidence} itself to make an inference of the class label. 

% his suggests that the wide range of different claims we generate as training data helps eliminat
% me of the annotation artifacts present in FEVER,leading to a more robust fact verification model

\begin{table}[htb]
  \small
  \begin{center}
      \begin{tabular}{|cc|ccc|}
      \hline
        \multicolumn{2}{|c|}{Dataset} & Supported & Refuted & NEI \\ \hline \hline
        \multirow{2}{*}{FEVER} & Train & 80,035 & 29,775 & 35,517 \\
        & Test & 6,666 & 6,666 & 6,666 \\
        \multicolumn{2}{|c|}{FEVER-Symmetric} & 710 & 710 & $-$ \\ \hline
        \multirow{2}{*}{\model} & Full & 176,370 & 360,924 & 258,452 \\ 
        & Filtered & 100,000 & 100,000 & 100,000 \\ \hline
      \end{tabular}
  \end{center}
  \caption{Basic statistics of the FEVER dataset and the dataset generated by \model. }
  % \caption{Basic statistics of the fact verification datasets . }
  \label{tbl:datasets}
\end{table}

% \begin{figure*}[htb]
% \centering
% \subfigure[FEVER-S/R]
% {
% 	\begin{minipage}[t]{0.31\linewidth}
% 	\centering
% 	\includegraphics[width=5.2cm]{acl-ijcnlp2021-templates/figures/fewshot_binary_roberta.png}
% 	\end{minipage}
% 	\label{fig:few_shot_SR}
% }
% \subfigure[FEVER-S/R/N]
% {
% 	\begin{minipage}[t]{0.31\linewidth}
% 	\centering
% 	\includegraphics[width=5.2cm]{acl-ijcnlp2021-templates/figures/fewshot_threeclass_roberta.png}
% 	\end{minipage}
% 	\label{fig:few_shot_SRNEI}
% }
% \subfigure[FEVER-Symmetric]
% {
% 	\begin{minipage}[t]{0.31\linewidth}
% 	\centering
% 	\includegraphics[width=5.2cm]{acl-ijcnlp2021-templates/figures/fewshot_symmetric_roberta.png}
% 	\end{minipage}
% 	\label{fig:few_shot_Symmetric2}
% }
% \vspace{-0.3cm}
% \caption{The few-shot learning experiment. The figure shows the F1 score on the three different settings for progressively larger training dataset sizes. Note the difference in scales for the Y-axes.}
% \vspace{-0.3cm}
% \label{fig:few_shot}
% \end{figure*}

\section{Model Implementation Details}
\label{app:baseline_implementation}

% We give implementation details for the fact verification models presented in the main result table. 

\paragraph{BERT-base and RoBERTa-large (S1, S2, U5, U6).}
We use the \texttt{bert-base-uncased} (110M parameters) and the \texttt{roberta-large} (355M parameters) model provided by HuggingFace library to implement the BERT model and the RoBERTa model, respectively. The model is fine-tuned with a batch size of $16$, learning rate of 1e-5 and for a total of $5$ epochs, where the epoch with the best performance is saved. 

\paragraph{GPT2 Perplexity (U2).}
To measure the perplexity, we use the HuggingFace implementation of the medium GPT-2 model (\texttt{gpt2-medium}, 345M parameters). We then rank the claims in the FEVER test set by their perplexity under the GPT-2 model. We then predict the label for each claim based on the assumption that misinformation has high perplexity. However, manually setting the perplexity threshold is difficult. Since the FEVER test set contains an equal number of claims for each class, we predict the claims in the top 1/3 of the ranking list as \texttt{refuted}, and the bottom 1/3 as \texttt{supported}. The rest claims are set as \texttt{NEI}. Therefore, the number of predicted labels for each class is also equal. 

\paragraph{MNLI-Transfer (U3).}
We use the HuggingFace – BERT base model (110M parameters) fine tuned on the Multi-Genre Natural Language Inference (MNLI) corpus\footnote{https://huggingface.co/textattack/bert-base-uncased-MNLI}, a crowd-sourced collection of 433K sentence pairs annotated with textual entailment information. We then directly apply this model for fact verification in the FEVER test set. The class label \texttt{entailment}, \texttt{contradiction}, and \texttt{neutral} in the NLI task is mapped to \texttt{supported}, \texttt{refuted}, and \texttt{NEI}, respectively, for the fact verification task. 

\paragraph{LM as Fact Checker (U4).}
Since there is no public available code for this model, we implement our own version following the settings described in~\citet{DBLP:LM_as_fact_checker}. We use HuggingFace's \texttt{bert-base} as the language model to predict the masked named entity, and use the NLI model described in U3 as the entailment model. 

% \section{Few-shot Learning Results}

% Figure~\ref{fig:few_shot} shows the few-shot learning curve for all three settings. Results on all three settings show that pretraining with claims generated by \model can greatly reduce the need of human annotation for training data. 

\end{document}